\documentclass{article}
\usepackage{spconf,amsmath,graphicx}
\usepackage{times}
\usepackage{url}
\usepackage{latexsym}
\usepackage{amsmath,amssymb,amsfonts}
\usepackage{algorithmic}
\usepackage{graphicx}
\usepackage{textcomp}
\usepackage{comment}
\usepackage{xcolor}
\usepackage{CJK}
\usepackage{color}
\usepackage{booktabs}
\usepackage{multirow}
\usepackage{multicol}
\usepackage{mathrsfs}

\title{Towards Zero-Shot Personalized Table-to-Text Generation with Contrastive Persona Distillation}

\name{Haolan Zhan$^{1,2,*}$, Xuming Lin$^{2}$, Shaobo Cui$^{3,*}$, Zhongzhou Zhao$^{2}$, Wei Zhou$^{2}$, Haiqing Chen$^{2}$\thanks{$^*$Work done when authors were at DAMO Academy, Alibaba.}}
\address{$^1$Monash University, Australia \quad $^2$DAMO Academy, Alibaba Group, China \quad $^3$EPFL, Switzerland}
\begin{document}
%
\maketitle
\begin{abstract}

Existing neural methods have shown great potentials towards generating informative text from structured tabular data as well as maintaining high content fidelity. However, few of them shed light on generating personalized expressions, which often requires well-aligned persona-table-text datasets that are difficult to obtain. To overcome these obstacles, we explore personalized table-to-text generation under a zero-shot setting, by assuming no well-aligned persona-table-text triples are required during training. To this end, we firstly collect a set of unpaired persona information and then propose a semi-supervised approach with contrastive persona distillation (S\textsuperscript{2}P-CPD) to generate personalized context.
Specifically, tabular data and persona information are firstly represented as latent variables separately. Then, we devise a latent space fusion technique to distill persona information into the table representation. Besides, a contrastive-based discriminator is employed to guarantee the style consistency between the generated context and its corresponding persona. Experimental results on two benchmarks demonstrate S\textsuperscript{2}P-CPD's ability on keeping both content fidelity and personalized expressions.


\end{abstract}

%

\section{Introduction}

Table-to-text generation, aiming at generating natural and informative context from structured tabular data~\cite{reiter1997building,puduppully2019data,zhao2020bridging,chang2021order}, has gained increasing attention for its tremendous value in many real applications such as advertising text generation on E-commerce~\cite{chen2019towards,shao2019long,zhan2020user} or headline generation on News recommendation~\cite{zhang2019outline,zhang2019multi}.
In these fields, content fidelity and personalized expressions are two main key factors. 
However, existing approaches~\cite{wang2020towards,gong2020enhancing} on table-to-text generation mostly pay their efforts to keep high content fidelity, but fall short of the personalized expressions. One primary reason is that the well-aligned persona-data-text triples are scarce in the real world, and collection of such pair-wised personalized dataset is usually labor-intensive and time-consuming. 




Previous work~\cite{ye2020iclr,lin2020imitate,zhan2021probing} tried to diversify the table-to-text generation system with content style by directly copying specific style phases from implicit or explicit exemplars.
While the success of these "hard" copy methods is indisputable, it’s inevitable for them to introduce noise data and do harm to the content fidelity.
In this work, we explore the personalized table-to-text generation under a zero-shot setting, where no well-aligned persona-table-text triples are required during training. Instead, our method will allow developers to diversify the system with persona characteristics from independent tabular data (e.g., table-text pairs collected from Taobao~\cite{chen2019towards} or Wikipedia~\cite{lebret2016wiki}) and persona information (e.g., user profiles).
Thus, it can greatly reduce the cost of building such systems and enhance the generalization.

\begin{figure}[!t]
\centering
\includegraphics[width=1.0\linewidth]{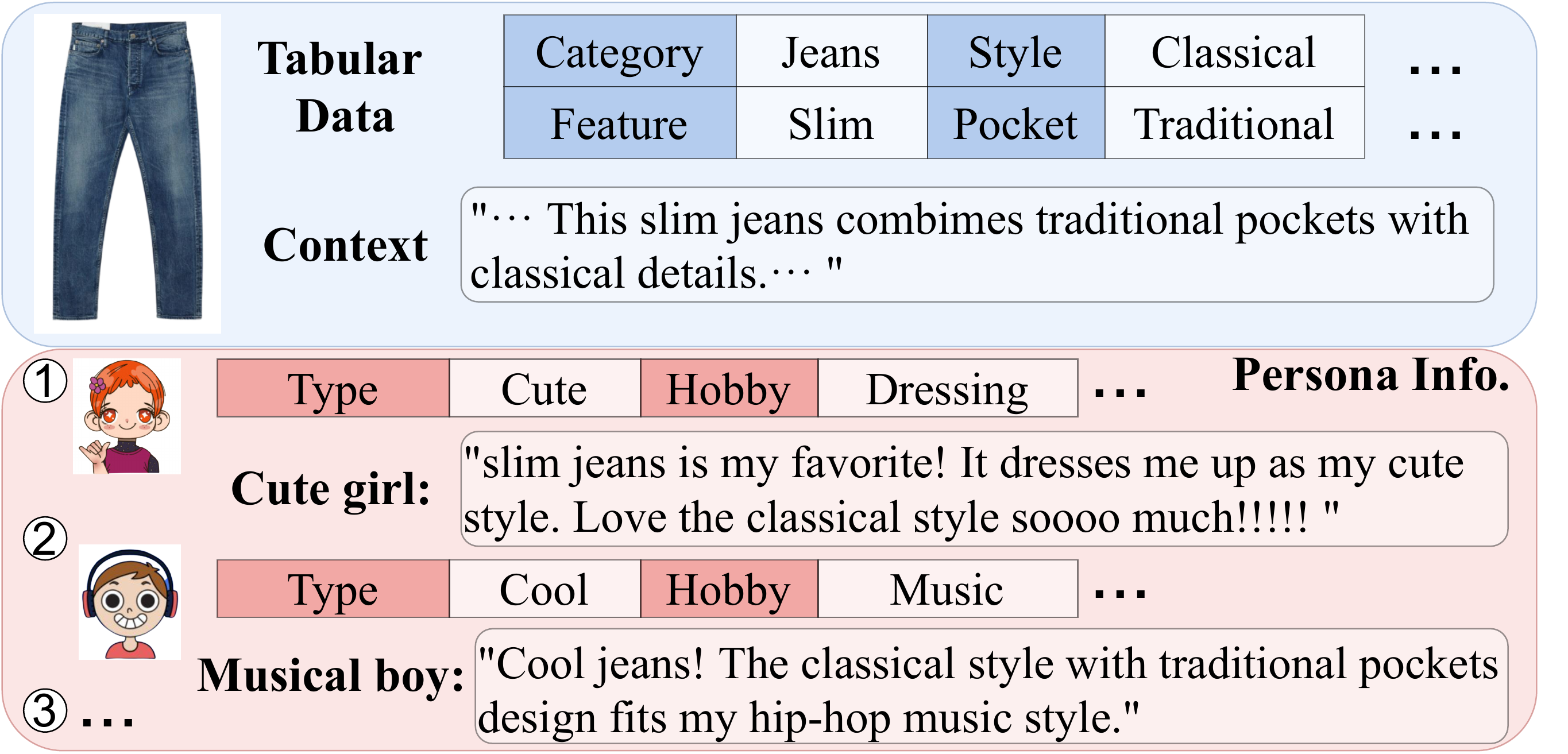}
\vspace{-5mm}
\caption{While conventional  table-to-text generation
methods mainly focus on keeping high content fidelity (Blue), personalized expressions (Red) are also important in real scenarios, e.g., E-commerce and News recommendation.}
\label{fig:intro}
\vspace{-3mm}
\end{figure}

To tackle the aforementioned obstacles and efficiently incorporate persona information into the generated context, we devise a novel semi-supervised framework with contrastive persona distillation (S\textsuperscript{2}P-CPD), containing persona distillation and contrastive-based discriminator modules. 
Specifically, we firstly utilize two auto-encoder modules to encode tabular data and persona profiles into intermediate latent variables respectively.
Then, a latent space fusion technique is devised to distill persona information into the table representation. Besides, a contrastive-based discriminator is employed to guarantee the personalized style consistency between the generated context and its corresponding persona.

We conduct our experiments on two table-to-text datasets, \textit{Taobao Advertising}~\cite{shao2019long} and \textit{WikiBio}~\cite{lebret2016wiki}. Besides, to make it appropriate to our task, we also collect two corresponding unpaired persona datasets from \textit{Hamlet} and \textit{Weibo} respectively.
Experimental results show our model is able to incorporate persona style into table-to-text generation, while also keep high content fidelity. 
Our contributions are threefold: 
(1) We firstly collect two sets of unpaired persona information, for better development of this task. (2) We propose a new semi-supervised method for personalized table-to-text generation by utilizing unpaired persona.
It's more practical in the real scenarios where pair-wised corpora are limited. 
(3) We conduct comprehensive experiments on two large-scale datasets to demonstrate the superiority of our model. 

\section{Methodology}

\subsection{Task Formulation}
For a given structured table data $\mathcal{X}$, when combined with different types of user profile $u$, we expect to generate different personalized context $y_u$. However, well-aligned ($\mathcal{X}, u, y_u$) triples are limited. 
Specifically, we have data-text pairs that are independent to persona information: $\mathcal{D}_{d} \equiv \{\left\langle \mathcal{X}_i, y_i \right\rangle \}^{D}_{i=1}$, in which $\mathcal{X}_i =\{X_{i}^{j}\}^{m}_{j=1}$ is a set of structured data and $X_{i}^{j} = \left({t^{j}, a^{j}}\right)$ consists of data type $t^{j}$ and its attribute value $a^{j}$ (as shown in Fig.~\ref{fig:intro}). $y_i$ is the ground-truth context sentence with no persona style. $u_{i}$ is the independent person information sentence.
The goal of unpaired personalized data-to-text is to generate personalized context under a zero-shot setting, where the training data-text pairs are \text{not} paired with persona label. More formally, 
The goal of our model is to learn to generate personalized context $y_u$  following $p(y_u|X,u)$, by not only considering the pairwise structured table data but also incorporating the unpaired persona information.

\begin{figure}[!t]
\centering
\includegraphics[width=1.0\linewidth]{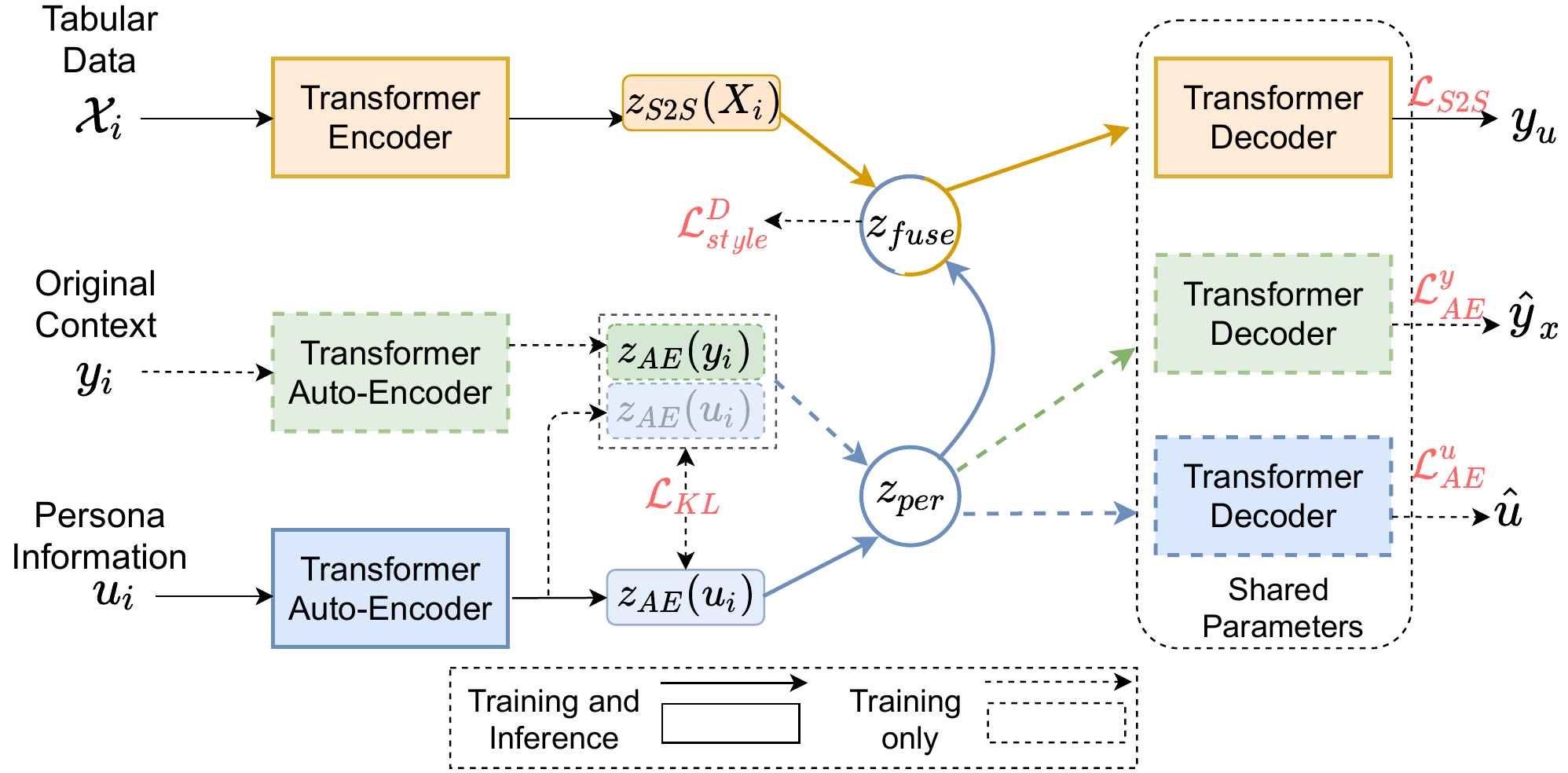}
\vspace{-5mm}
\caption{Overall framework architecture of S\textsuperscript{2}P-CPD.}
\label{fig:framework}
\vspace{-3mm}
\end{figure}



\subsection{S\textsuperscript{2}P-CPD Framework} \label{sec:model:overview}

\noindent \textbf{\textit{Overviews.}}
Fig.~\ref{fig:framework} presents our proposed semi-supervised framework with contrastive persona distillation (S\textsuperscript{2}P-CPD). This framework depicts the relationship among tabular data-text pair, unpaired persona information and their corresponding latent variables.
During training, S\textsuperscript{2}P-CPD framework consists of three main parts: 
(1) Encoder module 
(2) Persona distillation
(3) Contrastive-based discriminator. 
Encoder module contains Seq2seq (S2S) module Autoencoder (AE) module. S2S module is utilized to encode original tabular data $\mathcal{X}_i$ into a latent representation $z_{S2S}(X_i)$, and AE module is used to encode its aligned context $y_i$ and unpaired persona information $u_i$ into latent representation $z_{AE}(y_i)$ and $z_{AE}(u_i)$ respectively. Then, we utilize a latent space fusion technique to distill persona information into $z_{fuse}$. Besides, a contrastive-based discriminator is employed to guarantee the style consistency.
Both S2S and AE modules are initially parameterized by the Transformer-based model~\cite{vaswani2017attention}. We focus on the interaction among latent variables, and omit the details  inside encoder and decoder. 
We refer~\cite{vaswani2017attention} for more details about the internal architecture of Transformer model.

\noindent \textbf{\textit{Encoder for Table and Persona.}}
The encoder modules in autoencoders are used to encode $y_i$ and $u_i$ into latent representations $z_{AE}(y_i)$ and $z_{AE}(u_i)$ separately. Note that although the autoencoders for $y_i$ and $u_i$ have a shared decoder, their encoders are separate. 
The auto-encoder module empowers the model with the ability of generating personalized content, especially in the situation that strong supervision towards personalized context is absent.  It can help force the S\textsuperscript{2}P-CPD model to memorize the persona information.
Therefore, we combine $z_{AE}(y_i)$ and $z_{AE}(u_i)$, defined as follows:
\begin{equation*}
z_{per} = \lambda\cdot z_{AE}(y_i) + (1 - \lambda)\cdot z_{AE}(u_i),
\end{equation*}
where $\lambda$ is a hyper-parameter to balance these two latent representations for context and the persona sentence: $z_{AE}(y_i)$ and $z_{AE}(u_i)$. $z_{per}$ is used to reconstruct $y_i$ and $u_i$ during training. While during inference step, $z_{per}$ is constructed by $z_{AE}(u_i)$. The gap between training and inference is minimized by a KL distillation~\cite{hinton2015distilling} loss: 
\begin{equation*}
   \mathcal{L}_{KL}(\theta)= D_{KL}(p(z_{per}) || q(z_{per});\theta),
\end{equation*}
where $p(\cdot)$ and $q(\cdot)$ denote the prior and posterior representations for inference and training steps respectively. The auto-encoder objectives of $\mathcal{L}_{AE}^{y}$ and $\mathcal{L}_{AE}^{u}$ are cross-entropy loss

\noindent \textbf{\textit{Persona Distillation.}}
Our goal is to not only generate consistent context with the input tabular data but also to encourage our model to generate similar personalized style with the persona information. Therefore, we propose a latent representation fusion mechanism between $z_{S2S}(X_i)$ and $z_{per}$ to encourage the persona distillation into table latent representation, while $z_{S2S}(X_i)$ is encoded from S2S encoder module: 
\begin{equation*}
    z_{fuse} = \beta \cdot z_{S2S}(X_i) + (1 - \beta)\cdot 
    z_{per},
\end{equation*}
where $\beta$ is the hyper-parameter to balance these two latent representation. 
To better fuse the latent representation, we use $z_{AE}(y_i)$ as the intermediary. Specifically, we firstly fuse the $z_{AE}(u_i)$ and $z_{AE}(y_i)$ to get the $z_{per}$. Then we fuse the $z_{per}$ and $z_{S2S}(X_i)$ to get the $z_{fuse}$. Then, $z_{fuse}$ will be forwarded to the decoder phase to generate ${y}_{u}$, and used for contrastive-based discriminator module.



\noindent \textbf{\textit{Contrastive-based Discriminator.}}
We propose a contrastive-based discriminator to ensue the persona style consistency between the generated context and the given persona information. Nevertheless, a discriminator equipped with conventional MLE objective~\cite{shen2016minimum} would quickly saturate after the model learns the persona style difference.
Therefore, inspired by contrastive learning~\cite{hadsell2006dimensionality,gutmann2012noise}, we propose an advanced discriminator based on contrastive learning, which not only learns from positive instances but also from negative examples. 

Before applying our contrastive-based discriminator to S\textsuperscript{2}P-CPD model, we need to pre-train a baseline discriminative model which is capable to justify if a given sentence contains its corresponding persona style label. The pre-trained discriminative baseline model is denoted as $p_{b}(\cdot;\phi)$, whose value falls in $[0,1]$. The closer this value is to 1, the higher of the confidence. 


Intuitively, we advocate the use of contrastive learning to explicitly perceive the difference between the positive and negative samples. Specifically,  we utilize the pre-trained baseline model $p_{b}(·; \phi)$ to provide the target discriminator model $p_{m}(·; \theta)$ with a strong reference when contrasting the true persona label and the wrong one. Therefore, the target discriminator model is expected to give a higher confidence score $p(t_u|{z_{fuse}})$ on the persona style label $t_u$ for the true category, and a lower score for those wrong categories, compared to the value given by the pre-trained baseline model. We define the difference between $p_{b}(·; \phi)$ and $p_{m}(·; \theta)$ as:
\begin{equation*}
    \mathcal{D}((t_u,{z_{fuse}});\theta,\phi) = {\rm log}\frac{p_{m}(t_u|{z_{fuse}},\theta)}{p_{b}(t_u|{z_{fuse}},\phi)}.
\end{equation*}

For positive pair, we expect $\mathcal{D}((t_u,{z_{fuse}});\theta,\phi) > 0$, and for negative pair, we expect it less than 0. Therefore, we minimize the following objective:
\begin{align}
\footnotesize
    \nonumber \mathcal{L}^{D}_{style}(\theta; \phi) =
    &-\frac{1}{N}\sum {\rm log}\sigma(\mathcal{D}((t_u, {z_{fuse}})^{+};\theta,\phi))\\
   \nonumber  &-\frac{1}{N}\sum {\rm log}[1 - \sigma(\mathcal{D}((t_u,{z_{fuse}})^{-};\theta,\phi))],
\end{align}
where $\sigma(\cdot)$ is the sigmoid function. The given training pairs $(t_u,{z_{fuse}})$ can be used as positive samples $(t_u,{z_{fuse}})^{+}$. Negative samples $(t_u,{z_{fuse}})^{-}$, 
however, are obtained by negative sampling.





\noindent \textbf{\textit{Traning Objectives.}}
The final training objective is the combination of the aforementioned three parts:
\begin{equation*}
\tiny
    \mathcal{L}_{total} = \mathcal{L}^{D}_{style} + \mathcal{L}_{KL} + \mathcal{L}_{S2S} + (\mathcal{L}_{AE}^{y} + \mathcal{L}_{AE}^{u}),
\end{equation*}
where $\mathcal{L}_{S2S}$, $\mathcal{L}_{AE}^{y}$ and  $\mathcal{L}_{AE}^{u}$ are all cross-entropy loss for the S2S and AE modules respectively. $\mathcal{L}^{D}_{style}$ is the loss for the contrastive discriminator, and $\mathcal{L}_{KL}$ is for persona distillation.

\section{Datasets}

\begin{CJK*}{UTF8}{gbsn}



We use two table-to-text datasets: \textit{Taobao Advertising}~\cite{shao2019long} 
in Chinese,  and  \textit{WikiBio}~\cite{lebret2016wiki} 
in English. Plus, to make it appropriate to our framework, we also  collect two sets of unpaired persona profiles respectively. \textit{Taobao Advertising} is collected from a well-known Chinese E-commerce website Taobao\footnote{https://www.taobao.com}. 
This dataset mainly focuses on the category of clothes and shoes. This dataset is designed to generate product descriptions from a set of structured product attributes (e.g. "category", "feature" for the jeans in Fig.~\ref{fig:intro}). It contains 114K/3K/3K cases for train/dev/test. For its corresponding unpaired persona information, we collect a set of perosna profiles from Weibo\footnote{{https://www.weibo.com}},
a popular Chinese social platform. The \textit{Weibo} dataset contains 4,653 instances. For each instance, persona information contains persona label, sex, age and reviews. There are 5 different persona labels (categories) including: "体育迷 (sport fan)", "音乐控 (music amateur)", "IT达人 (IT fancier)", "电影范 (movie lover)" and "吐槽粉 (critic)". These labels will be used for the training of the pre-trained baseline and contrastive-based discriminator. 

Besides, the English dataset is \textit{WikiBio}, a classical benchmark in the table-to-text generation task. It aims to translate structured biography information from the Wikipedia into a paragraph of description. It contains 582K/72K/72K on the train/dev/test. Besides, we collect the corresponding unpaired dataset from  \textit{Hamlet}\footnote{\textit{Hamlets} is one of the famous Shakespeare Tragedies.}, one of the most famous Tragedies of Shakespeare. For the \textit{Hamlet} dataset, persona information according to their names, includes Hamlet (Prince), Claudius (King), Queen, etc. For each person, the persona information will include their ages, speaking words. In total, we collect 1,548 instances from the most five frequent characters. Their names will be taken as their persona labels (categories).

\end{CJK*}

\begin{CJK*}{UTF8}{gbsn}

\begin{table*}[!ht]
\centering
\scriptsize
\caption{\label{tb:example} Case study on the \textit{Taobao Advertising + Weibo} dataset.}
\newcommand{\tabincell}[2]{\begin{tabular}{@{}#1@{}}#2\end{tabular}}
\begin{tabular}{ll} 
\toprule
    Tabular Data & \tabincell{l}{\textbf{类型}: 上衣, \textbf{版型}: 宽松, \textbf{版型}: 显瘦, \textbf{图案}: 线条, \textbf{衣袖型}: 泡泡袖, \textbf{款式}: 抽绳 \\ \textbf{Type}: Coat, \textbf{Pattern}: Loose, \textbf{Pattern}: Slim, \textbf{Layout}: Line, \textbf{Sleeves}: Bubble, \textbf{Layout}: Drawstring} \\
   \hline
    Paired Ground Truth &\tabincell{l}{ 一款很 有设计感衬衫, 采用了宽松的版型剪裁, 衣身采用了泡泡袖 设计, 修饰手臂线条, 抽绳设计方便穿脱。\\(A type of fashionable coat, with loose pattern. This coat takes the special bubble design to slim the line of arms.  \\The layout of drawstring is friendly to use.)}\\
   \hline
   Upaired Persona & \tabincell{l}{性别: 女, 年龄:18, 类别: 音乐控, 评论: 妈呀！萌的我一脸，萌萌哒的千与千寻，这个版本很好听呢！} \\
   Profiles & \tabincell{l}{\textbf{Sex}: female, \textbf{Age}: 18, \textbf{Type}: music fan, \textbf{Review}: My god! It's so cute! A very cute version of Spirited Away! This version is very good. } \\\hline
   {Sty-Im} &\tabincell{l}{这款上衣十分的宽松，显瘦的版型萌萌哒， 泡泡袖的设计很好听呢！ 抽绳的设计，方便穿脱。(This coat is very loose, the \\slim pattern is very cute, the design of bubble sounds really good!. The drawstring style is easy for user to take off.)} \\ \hline
   \textbf{Ours} &\tabincell{l}{这款衬衫采用了宽松的版型， 遮肉显瘦的效果也是棒棒哒。可爱的泡泡袖设计, 尽显甜美俏皮感哦。腰部的抽绳设计，尽显\\纤细好身材呢！ (This coat takes a loose pattern, It's very very good to show your slim body. The very cute bubble design is able\\ to show your sweet and beauty! The drawstring design over your waist will show your good body without hesitate. Wink~)}\\
  \bottomrule
\end{tabular}
\vspace{-1mm}
\end{table*}

\end{CJK*}

\begin{table}[!t]
\footnotesize
\centering
\caption{Automatic evaluation results on (a) {\it Taobao Advertising + Weibo } and (b) {\it WikiBio + Hamlets}. 
}
\resizebox{0.45\textwidth}{30mm}{
\begin{tabular}{c|l|cccc}
\toprule[1pt]
   ~ & \multirow{2}{*}{Model} & \multicolumn{4}{c}{\textit{Content Fidelity}}  \\
   ~ & & ACC & PPL & BLEU & ROUGE-L  \\
  \hline
  \multirow{7}{*}{(a)} & S2SA-Copy & 68.07 & 133.26 & 7.85 &  15.79  \\
  & Trans-Copy & 72.51 & 73.64 & 9.63 & 17.22   \\
   & PHVM & \textbf{86.30} & 41.80 & 12.75 & 22.91  \\
   & Sty-Im & 79.62 & 82.61 & 11.49 & 20.55   \\ \cline{2-6}
   & S\textsuperscript{2}P-CPD \textit{w/o Per.} & 82.61 & 40.63 & 11.93 & 22.16   \\
   & S\textsuperscript{2}P-CPD \textit{w/o CD} & 81.55 & 44.07 & 12.54 & 21.69   \\
   & S\textsuperscript{2}P-CPD & 82.49 & \textbf{36.28} & \textbf{13.20} &\textbf{24.67}   \\\hline\hline
   \multirow{7}{*}{(b)} & S2SA-Copy & 77.39 & 50.81 & 38.46  & 39.21 \\
   & Trans-Copy & 76.34 & 49.27 & 39.10  & 41.06  \\
   & PHVM & 86.02 & 28.94 & \textbf{43.87} & 45.25 \\
    & Sty-Im &83.25 & 31.32 & 41.38 &43.18 \\\cline{2-6} 
    & S\textsuperscript{2}P-CPD \textit{w/o Per.} & 84.33 & 27.71 & 41.82  & 43.50  \\
   & S\textsuperscript{2}P-CPD \textit{w/o CD} & 85.47 & 24.88 & 40.67  & 44.61  \\
   & S\textsuperscript{2}P-CPD & \textbf{87.64} & \textbf{21.75} & 42.96  &\textbf{45.39} \\
   
 \bottomrule[1pt]
\end{tabular}
}
\label{tab_eval}
\vspace{-2mm}
\end{table}

\section{Experiments}

\begin{CJK*}{UTF8}{gbsn}


\noindent \textbf{\textit{Baselines.}}
We compare our model with four baseline models including: (1) \textbf{S2SA-Copy} is a LSTM-based model with copy and attention mechanism; (2) \textbf{Trans-Copy} is a vanilla Transformer with copy mechanism; (3) \textbf{PHVM} is a planning-based hierarchical VAE model~\cite{shao2019long}; (4) \textbf{Sty-Im} is a style imitation model with Transformer for data-to-text generation task~\cite{lin2020imitate}.
Besides, we also evaluate two ablation models: S\textsuperscript{2}P-CPD \textit{w/o Per.} removes the persona distillation, and S\textsuperscript{2}P-CPD \textit{w/o CD} removes the contrastive-based discriminator.

\noindent \textbf{\textit{Hyper-Parameters.}}
 For the involved experimental models, the hidden units of all transformer-based models are set as 512 and the feed-forward hidden size is set as 1,024. The beam search size is set as 5 and length penalty as $\alpha$ = 0.4. The initial learning rate is set to 0.001. The $\beta_1$ = 0.9 and $\beta_2$ = 0.998 are used for gradient optimization. We also apply warm-up trick over the first 8,000 steps, and decay as in \cite{vaswani2017attention}.

\noindent \textbf{\textit{Evaluation Metrics.}}
We evaluate the performance of our model and baselines from three aspects: (1) Content fidelity is evaluated by keywords accuracy (ACC) on the attribute data, BLEU~\cite{papineni2002bleu} and ROUGE-L~\cite{lin-2004-rouge}. (2) Fluency is evaluated by perplexity (PPL)~\cite{chen1998evaluation}. (3) Persona engagement is evaluated by human annotators, and we randomly select 200 cases from the test set for human evaluation.

\noindent \textbf{\textit{Automatic Evaluation Results.}}  
Table~\ref{tab_eval} shows the evaluation results on the two datasets.
As we can see, our proposed approach achieves a competitive performance, while comparing the state-of-the-art baseline model on the traditional data-to-text generation task, such as PHVM. Besides, in terms of the persona engagement, we carry out a human evaluation on the generated context, and find that our approach achieves better performances than Sty-Im. Although Sty-Im could reach a relatively higher performance than other baselines without persona incorporation, it still introduces noise and irrelevant information. Besides, the results on two ablation models also demonstrate the effectiveness of S\textsuperscript{2}P-CPD.
Finally, we also carry out the  statistically significant test with the $p < 0.01$.

 \begin{table}[!t]
\centering
\footnotesize
\caption{Human evaluation between S\textsuperscript{2}P-CPD and other baselines on the (a) {\it Taobao Advertising + Weibo } dataset.}
\begin{tabular}{c|c|ccc|c}
  \toprule[1pt] 
  \multirow{2}{0.9cm}{Dataset} & \multirow{2}{0.8cm}{Model} & \multicolumn{3}{c|}{S\textsuperscript{2}P-CPD vs.} & \multirow{2}{0.8cm}{kappa}\\ 
\cline{3-5} 
  ~ & ~ & Win & Loss & Tie & ~ \\
  \hline
  \multirow{4}{*}{(a)} & S2SA-Copy  & 63\% & 8\% & 29\% & 0.603 \\
  ~ & Trans-Copy  & 49\% & 18\% & 33\% & 0.558 \\
  ~ & PHVM  & 53\% & 27\% & 20\% & 0.526 \\
  ~ & Sty-Im & 42\% & 30\% & 28\% & 0.492  \\ 
  \bottomrule[1pt]
\end{tabular}
\label{tab:human}
\vspace{-2mm}
\end{table}

\noindent\textbf{\textit{Human Evaluation Results.}}  These results are shown in Table \ref{tab:human}. We observe that S\textsuperscript{2}P-CPD outperforms all baseline models on the {\it Taobao Advertising + Weibo }, where results on {\it News + Hamlets} maintains the consistent trend. Specifically, the percentage of ``win'' is always larger than that of ``loss''. Compared with Trans-Copy, PHVM and Sty-Im, S\textsuperscript{2}P-CPD achieves preference gains (win subtracts loss) with 31\%, 26\% and 12\%, respectively. We check responses generated by our model with ``win'' and find that they are more relevant to contextual utterances. The kappa scores~\cite{fleiss1971measuring} indicate that annotators come to a ``moderate agreement'' on judgement.

\noindent\textbf{\textit{Case Study.}} \quad 
We present a case study on the \textit{Taobao Advertising + Weibo} dataset. We observe that our model outperforms the most relevant baseline Sty-Im by not only engage the persona style but also keep high content fidelity with original tabular data. For example, although Sty-Im could present some personalized expressions by copying from persona information, it also introduces some noise messages ("好听 (sounds good)"), which is counterfactual and irrelevant with original data. S\textsuperscript{2}P-CPD is able to incorporate personalized expressions appropriately and also keep the faithfulness. This is because we take a "soft" latent fusion and persona distillation method to incorporate persona  style.

\end{CJK*}
\section{Conclusion}

In this paper, we study the problem of zero-shot table-to-text generation with personalized expressions . We propose a semi-supervised learning framework, in which persona distillation and contrastive-based discriminator are employed to endow the system with personalized expressions. Experimental results show that our approach achieves a good balance between content fidelity and persona expression, and is flexible to adapt different personal profiles. 

\section*{Acknowledgement}

This work was supported by Alibaba through Alibaba Research Intern Program. Thanks all reviewers for comments.

\vfill\pagebreak

\bibliographystyle{IEEEbib}
\bibliography{icassp2022ref}






\end{document}